%% file: paper_lrec.tex
\documentclass[10pt, a4paper]{article}

\usepackage{amsmath}
\usepackage[linesnumbered,ruled,vlined]{algorithm2e}

\usepackage[switch]{lineno}   

\usepackage{listings}
\usepackage[T1]{fontenc}

\usepackage[utf8]{inputenc}

\usepackage{microtype}

\usepackage{graphicx}
\usepackage{color, xspace}

\newcommand{\textttnew}[1]{\lstinline!#1!}
  \usepackage{amsthm}

 \theoremstyle{remark}
  \newtheorem{remark}{Remark}

\usepackage[final]{lrec2026} 

\title{Node-Level Uncertainty Estimation in LLM-Generated SQL}

\name{Hilaf Hasson, Ruocheng Guo} 

\address{Intuit AI Research\\
         \{hilaf\_hasson, ruocheng\_guo\}@intuit.com\\}

\abstract{
We present a practical framework for detecting errors in LLM-generated SQL by estimating uncertainty at the level of individual nodes in the query’s abstract syntax tree (AST). Our approach proceeds in two stages. First, we introduce a semantically aware labeling algorithm that, given a generated SQL and a gold reference, assigns node-level correctness without over-penalizing structural containers or alias variation. Second, we represent each node with a rich set of schema-aware and lexical features - capturing identifier validity, alias resolution, type compatibility, ambiguity in scope, and typo signals - and train a supervised classifier to predict per-node error probabilities. We interpret these probabilities as calibrated uncertainty, enabling fine-grained diagnostics that pinpoint exactly where a query is likely to be wrong. Across multiple databases and datasets, our method substantially outperforms token log-probabilities: average AUC improves by +27.44\% while maintaining robustness under cross-database evaluation. Beyond serving as an accuracy signal, node-level uncertainty supports targeted repair, human-in-the-loop review, and downstream selective execution. Together, these results establish node-centric, semantically grounded uncertainty estimation as a strong and interpretable alternative to aggregate sequence-level confidence measures.
\\ \newline  }

\begin{document}

\maketitleabstract

\section{Introduction}
With the advent of large language models (LLMs) \citep{achiam2023gpt, brown2020language, hoffmann2022training, guo2025deepseek}, recent advancements in Text-to-SQL have been largely driven by solutions that leverage LLMs to generate SQL queries, which now dominate benchmark leaderboards such as BIRD \citep{li2023can}, Spider 1.0 \citep{yu2018spider}, and Spider 2.0 \citep{lei2024spider}. (See \citealp{deng2025reforce, gao2023text, shkapenyuk2025automatic}.)

Errors in SQL generation—or, more broadly, in structured output generation—can be categorized into two main types: syntactic errors, which involve violations that make the SQL query unable to run; and semantic errors, often referred to as schema-linking errors in the context of SQL, which result in a SQL query that can be executed, but misinterprets the meaning of the question, including selecting the wrong table, column, or value. For syntactic errors, rule-based solutions abound, with \citealp{shen2025study} being salient among them. For semantic errors, following the general principle articulated in \citealp{kuhn2023semantic} that ``very uncertain generations should be less likely to be correct'', many state-of-the-art generative Text-to-SQL approaches incorporate voting mechanism (e.g., \citealp{deng2025reforce, shkapenyuk2025automatic}).

However, \citealp{xiong2023can} observed that LLMs can be overconfident in their responses, indicating that supervised uncertainty estimation could offer advantages when sufficient data is available. Since SQL generation is open-ended, we focus our approach on supervised node-level uncertainty quantification estimation. To that end, we begin by introducing the first semantically-aware labeling algorithm for correctness of nodes in pairs of generated and ground truth SQL -- a priori an ambiguous task (See Section \ref{ssec:ground_truth}). See Appendix ~\ref{app:examples} for how our design choices make the node-level correctness labeling agree with intuition in 13 cases we used while prototyping, while node-level hashing (the only alternative in related works) does not. We then create a training dataset with carefully designed, schema-informed, features and labels at the node level. (See Section \ref{ssec:featurization}.) Finally, we train a gradient-boosted tree model on this dataset to estimate node-level uncertainty. (See Section \ref{experiments}.) In particular, we interpret the term ``uncertainty'' as calibrated predictive probability of error.

Our experiments show conclusively that our proposed approach vastly outperforms the arithmetic mean of the logarithm of token probabilities from the LLM which generated the node as a predictor of error, suggesting the aforementioned principle in \citealp{kuhn2023semantic} can be significantly improved upon in the case of node-level uncertainty quantification.

More precisely, our contributions are as follows:

\begin{enumerate}
    \item In Section \ref{ssec:ground_truth}, given a generated SQL query and a ground truth SQL query, we give an algorithm for labeling what it means for a single node in the AST of a generated SQL to be correct (See Algorithm \ref{alg:labeling_highlevel}). See Appendix~\ref{app:examples}, which showcases how the design choices made the algorithm agree with intuition in 13 specific cases that we had used while prototyping.
    \item In Section \ref{ssec:featurization}, we outline our choice of 72 node-level schema-aware features.
    \item In Section \ref{experiments}, results for 3 experiments on the BIRD dev dataset show significant improvement over the baseline using log probabilities. In our first experiment, the training queries use the same databases from the BIRD dev dataset as those in the test set. In the second, the training queries come from different databases than those in the test set, demonstrating the model's ability to generalize to unseen databases. All 3 models outperform log probabilities significantly, but we observe that the more the training data was in-distribution (Experiment 1 > Experiment 2 > Experiment 3) the better the performance is. 
\end{enumerate}

\section{Related Work}

Most Text-to-SQL literature treats uncertainty quantification as an integral part of an overarching solution, which is not analyzed separately. For example \citealp{xie2025opensearch} employs consistency alignment, self-consistency voting, and error correction to mitigate hallucinations and variability, but does not quantify uncertainty outright.

The recent work \citep{somov2025confidence} introduces entropy-based confidence estimates with selective classifiers to enable query rejection under distribution shift. Complementing this, \citealp{liu2025calibrating} propose multivariate Platt scaling leveraging sub-clause frequency statistics across sampled SQL generations.

AST-based comparison has also been explored for SQL equivalence checking. The \texttt{sqlgpt-parser} package labels correctness at the \emph{query level}, treating two SQL queries as ``equal'' only if their ASTs are isomorphic trees with identical node types and attribute values at every position, matching nodes element-by-element in order and ignoring only line and position metadata \citep{sqlgptparser}. This approach does not include alias-to-base-table normalization, treatment of qualified vs. unqualified columns, handling of symmetric or anti-symmetric operators, or reconciliation between contextual and global alignment when structure differs.

Another line of work, PPOCoder \citep{shojaee2023execution} employs AST and data-flow graph (DFG) overlaps as sequence-level reward components in a reinforcement-learning framework for code generation. It does not define SQL-specific node equivalence, produce per-node correctness labels, or estimate calibrated node-level uncertainty.

A more recent work \citep{anonymous2024coarse}, is the first to introduce node-level uncertainty estimation in Text-to-SQL by training a transformer model (codellama 7B) that takes the original question, the schema, and the generated SQL, and outputs a confidence score per node. This differs from the our work in 2 significant ways: The first is that their solution is a black-box model over serialized input. Namely, the question, schema, and serialized generated SQL are all fed in as a single text, and model is expected to learn their structure implicitly. This is contrast to the current work, where nodes are featurized explicitly -- allowing for a more explainable model, and allowing to choose informative features. Second, their definition of ground truth is a very high level heuristic that is not semantically grounded: for each node in the generated sql they hash its subtree using a relational algebra tree (RAT); then they attempt to find a node with the same hash in the ground truth query. This causes errors to propagate upwards: when a child node is incorrect, all its ancestors are marked incorrect as well, even when their structure and logic are otherwise valid.

For example, with the ground truth SQL:
\begin{lstlisting}
SELECT department.name, SUM(employee.salary)
FROM department
JOIN employee ON department.id = employee.department_id
GROUP BY department.name
\end{lstlisting}

And model-generated SQL is:
\begin{lstlisting}
SELECT department.name, SUM(employee.wage)
FROM department
JOIN employee ON department.id = employee.department_id
GROUP BY department.name
\end{lstlisting}

The only error in this instance is the incorrect use of \texttt{employee.wage} in place of the correct column \texttt{employee.salary}. However, the RAT hashing mechanism compares entire subtrees. Since the subtree rooted at \textttnew{SUM(employee.wage)} does not match the one rooted at \textttnew{SUM(employee.salary)}, the \textttnew{SUM} node receives a different hash and is marked incorrect. There is also no explicit accounting of alias resolution, handling of symmetry and anti-symmetry, and so on; which causes equivalent statements to be labeled as incorrect.

\section{Proposed Method}
\label{sec:method}

Our framework identifies SQL errors by reframing query validation as a fine-grained classification problem at the level of individual query components, i.e., nodes in the AST of each SQL query. Any given SQL query is first parsed into an AST, a standard hierarchical representation of its structure. We then use a Gradient Boosted Decision Tree (GBDT) model to predict the probability that each node in the AST is erroneous. This node-level probability serves as a direct measure of uncertainty, pinpointing the specific parts of the query that are likely incorrect.

\subsection{Ground Truth Generation}
\label{ssec:ground_truth}

To train a robust supervised learning model for SQL error detection, we construct a dataset of machine-generated SQL queries annotated with fine-grained, node-level error labels. Each generated query is paired with a gold-standard, human-verified reference query. Labels are assigned through recursive comparison of their abstract syntax trees (ASTs) with ground truth, using a context-preserving, semantic alignment strategy.

\vspace{0.5em}
\textbf{Node Alignment and Labeling}
The labeling process performs a top‑down traversal over the generated AST. For each pair ($n_{\text{gen}}$, $n_{\text{gold}}$), we first test semantic recursive equivalence; which will be defined clearly below. If true, we mark all nodes in the subtree rooted at $n_{\text{gen}}$ as correct and stop recursing under this pair. Otherwise, we mark $n_{\text{gen}}$ as error and recurse on every child pair ($c_{\text{gen}}$, $c_{\text{gold}}$) formed by the children of $n_{\text{gen}}$ and the children of $n_{\text{gold}}$ (order‑invariant, cartesian). After this traversal, we suppress blame for structural containers (\textttnew{SELECT}, \textttnew{FROM}, \textttnew{WHERE}, \textttnew{GROUP BY}, \textttnew{HAVING}, \textttnew{=}, $\neq$) when only their children differ. 

This traversal continues recursively through the structure of the generated AST, with alignment decisions based only on local subtree comparisons; the gold AST is not searched globally during this phase. As a result, some semantically correct nodes may remain labeled as errors due to greedy or structurally mismatched alignment decisions. To further improve the labeling heuristic, we do a final, context‑agnostic pass that globally re‑evaluates nodes still labeled as errors—trading structural precision for increased recall. The precise algorithm is spelled out in Algorithm \ref{alg:labeling_highlevel}; Appendix ~\ref{app:examples} illustrates, across 13 cases, how these design choices align labels with intuitive judgments. 

\vspace{0.5em}
\textbf{Semantic Equivalence Criteria}

Two nodes $n_{\text{gen}}$ and $n_{\text{gold}}$ are considered semantically equivalent if the following conditions hold:

\begin{enumerate}
\item Type Errors:
\begin{enumerate}
    \item The node types are identical, \emph{or} they are a known operator pair (e.g., \textttnew{>} vs.\ \textttnew{<}) that permits equivalence under operand swapping.
    \item  If the nodes are binary operators (\textttnew{=}, \textttnew{>}, etc.), they are equivalent if:
\begin{itemize}
\item both children match in order, or
\item the operator is symmetric or anti-symmetric and the children match in reverse order.
\end{itemize}

\end{enumerate}
\item Content Errors:
\begin{enumerate}
    \item If the nodes are \textttnew{Column} expressions, their unqualified names must match exactly; qualifiers (aliases or table names) must either match directly, or resolve to the same base table via an alias-to-table map extracted from each AST.
\item If the nodes are literals, identifiers, or base tables, their content (e.g., string or numeric value) must match exactly.
\end{enumerate}

\end{enumerate}
\begin{remark}
To reduce over-labeling, the following heuristics are applied:
\begin{itemize}
    \item Nodes like \textttnew{SELECT}, \textttnew{FROM}, \textttnew{WHERE}, \textttnew{GROUP BY}, and comparison operators (\textttnew{=}, \textttnew{!=}) are not blamed if only their children differ.
    \item Structural nodes like \textttnew{JOIN} and \textttnew{LIMIT} \emph{are} blamed if they are incorrect or inserted.
    \end{itemize}
\end{remark}

\vspace{0.5em}
\textbf{Illustrative Example}

\begin{itemize}
    \item \textbf{Generated SQL:} \textttnew{SELECT name FROM artists ORDER BY name}
    \item \textbf{Gold SQL:} \hspace{1.1em} \textttnew{SELECT name FROM artist WHERE name = 'A'}
\end{itemize}

Here, the labeler would output:
\begin{itemize}
    \item An error for the table name \textttnew{artists} (vs.\ \textttnew{artist}),
    \item Errors for the additional \textttnew{ORDER BY} clause and its child nodes,
    \item Since the goal of the algorithm is only to label nodes in the generated SQL are labeled as either correct or incorrect, the missing \textttnew{WHERE} does not arise in the labeling,
    \item The \textttnew{name} in \textttnew{name = 'A'} will be labeled as incorrect in the first-pass traversal across generated and ground truth ASTs, but will be relabeled as correct in the final context-agnostic pass.
\end{itemize}
\begin{remark}
    Omissions are not penalized because they cannot be assigned node-level probabilities. Query-level completeness metrics (e.g., clause coverage or overlap with the gold query) is out of scope for this paper, but could be combined in future work.
\end{remark}
\vspace{0.5em}

We now proceed to spell out some details of the subtleties in the definition of Content Errors in the Semantic Equivalence Criteria.

\paragraph{Content Error Details}

Identifiers, literals, and function names must match exactly. Otherwise, it is a content error.

\begin{itemize}
    \item \textbf{Identifiers:} \textttnew{artists} vs.\ \textttnew{artist}
    \item \textbf{Literals:} \textttnew{1} vs.\ \textttnew{2}
    \item \textbf{Functions:} \textttnew{MAX()} vs.\ \textttnew{SUM()}
\end{itemize}
This includes mismatches in base table names and literal constants. 

Table nodes are considered correct iff their corresponding identifiers are correct after an alias-to-table mapping. Column nodes may include explicit qualifiers such as table names or aliases (e.g., \textttnew{a.name} vs.\ \textttnew{name}); these are treated as semantically equivalent when:
\begin{itemize}
    \item The column name is identical, and
    \item The qualifiers (if any) either match, or resolve to the same base table via alias mapping.
\end{itemize}

Alias names themselves are not required to match, and will not trigger content errors. A mapping from alias $\rightarrow$ base table is computed for both ASTs and used to normalize qualified comparisons.

\vspace{0.5em}

\paragraph{Post-Processing}
A final pass reclassifies any node that matches an equivalent node elsewhere in the gold tree, to correct for greedy alignment mismatches. (Note that the first, contextual, pass is not redundant because being contextually aware leads to more ``correct'' labels.)

\vspace{0.5em}
To show why the final pass is required consider the ground truth \textttnew{SELECT name FROM artists} and generated SQL \textttnew{SELECT name FROM artist}. Because the tree diverges at the level of the table, the first, contextual, pass of Algorithm \ref{alg:labeling_highlevel} would have labeled \textttnew{name} as incorrect, which does not agree with intuition.
\begin{algorithm}[!t]
\caption{Labeling Generated AST via Alignment with Gold AST}
\label{alg:labeling_highlevel}
\KwIn{Generated AST $T_{\text{gen}}$, Gold AST $T_{\text{gold}}$}
\KwOut{Label map $\mathcal{L}$ assigning 0 (correct) or 1 (error) to each node in $T_{\text{gen}}$}

Initialize $\mathcal{L}$ by labeling all nodes in $T_{\text{gen}}$ as 1\;

Build alias maps from both ASTs to normalize qualifiers\;

\SetKwFunction{F}{AlignAndLabel}
\SetKwFunction{Equiv}{AreRecursivelyEquivalent}

\textbf{function} \F{$n_{\text{gen}},\ n_{\text{gold}}$} \textbf{is} \\
\Indp
  \If{\Equiv{$n_{\text{gen}},\ n_{\text{gold}}$}}{
    Mark all nodes in the subtree rooted at $n_{\text{gen}}$ as correct in $\mathcal{L}$; \Return;
  }

  Mark $n_{\text{gen}}$ as error in $\mathcal{L}$\;

  \ForEach{child $c_{\text{gen}}$ of $n_{\text{gen}}$}{
    \ForEach{$c_{\text{gold}}$ in children of $n_{\text{gold}}$}{
    \F{$c_{\text{gen}},\ c_{\text{gold}}$}
    }
  }
\Indm
\textbf{end function} \\[1ex]

\F{root of $T_{\text{gen}}$, root of $T_{\text{gold}}$}\;

Suppress blame for structural nodes if only their children are labeled as errors\;

\ForEach{node $n_{\text{gen}}$ in $T_{\text{gen}}$ still labeled as error}{
  \If{$n_{\text{gen}}$ matches any $n_{\text{gold}}$ in $T_{\text{gold}}$ under \Equiv{}}{
    Relabel $n_{\text{gen}}$ as correct;
  }
}

\Return{$\mathcal{L}$}\;
\end{algorithm}
In Algorithm \ref{alg:labeling_highlevel} the function \textttnew{AreRecursivelyEquivalent} refers to verifying that the nodes are semantically equivalent in the sense of Section \ref{ssec:ground_truth}, that their children have a 1-1 correspondence of semantically equivalent nodes, and so on recursively. 

See Appendix ~\ref{app:examples} to see how 5 out of the 13 unit tests we include there for node-level correctness labeling to agree with intuition would have failed if not for the post-processing.

\subsection{Featurization}
\label{ssec:featurization}

The node classification model does not see the gold query; its ability to predict the ground truth label for a given node is derived entirely from the feature vector associated with a given node. Our featurization process is provides a rich, evidence-based description of each node, capturing its structure, semantic validity, and context. Features are grouped into several logical categories.

\input{table1}

\paragraph{1. Base Structural Features}
This foundational group describes the node's identity and position within the AST. It includes the node's syntactic type (e.g., \textttnew{Column}, \textttnew{Join}), its depth relative to the root, and the type of its immediate parent.

\paragraph{2. Schema Consistency Features}
This is a critical feature set for detecting content errors. It validates a node's content against the target database schema, using a scope resolution module that correctly interprets table aliases and nested queries. Key features include:
\begin{itemize}
    \item \textbf{Identifier Validity:} A binary feature checking if a table or column name exists in the schema. For an erroneous node like the \textttnew{Table} \textttnew{artists}, this feature provides a powerful error signal.
    \item \textbf{Qualifier Scope Validity:} For a qualified column like \textttnew{T.name}, this checks if the qualifier \textttnew{T} is a valid table or alias within the current scope of the query.
    \item \textbf{Column Ambiguity:} A feature that detects if an unqualified column name could refer to columns in multiple tables that are currently in scope, a common source of error in queries with multiple joins.
\end{itemize}

\paragraph{3. Lexical and Typo-Detection Features}
This group helps the model identify errors in identifiers that are syntactically well-formed but semantically incorrect (i.e., typos). These features are primarily for \textttnew{Identifier} and \textttnew{Column} nodes.
\begin{itemize}
    \item \textbf{Levenshtein Distance:} If an identifier is not found in the schema, we compute its Levenshtein (edit) distance to the closest valid identifier (table or column) in the entire schema. A small distance is a strong indicator of a typo.
    \item \textbf{Name Patterns:} A series of binary features capturing lexical patterns in an identifier's name, such as whether it contains numbers, underscores, or is written in `ALL\_CAPS' or `mixedCase'.
\end{itemize}

\paragraph{4. Contextual and Heuristic Features}
This broad category captures common SQL error patterns that depend on a node's relationship with other nodes, helping the model to infer structural and type-based errors.
\begin{itemize}
    \item \textbf{Aggregate Context:} A feature specifically designed to detect one of the most common SQL errors: the presence of a non-aggregated column in a \textttnew{SELECT} list that also contains an aggregate function (e.g., \textttnew{SUM()}) but lacks a corresponding \textttnew{GROUP BY} clause.
    \item \textbf{Data Type Compatibility:} For binary operations (\textttnew{=}, \textttnew{>}, etc.), we resolve the data types of the left and right operands and include a feature representing their compatibility (e.g., \textttnew{NUMERIC} vs. \textttnew{STRING}).
    \item \textbf{Operator-Specific Features:} For certain node types, we add highly specialized features. For a \textttnew{Like} node, we extract features about its pattern (e.g., presence of wildcards, pattern length). For an \textttnew{In} node, we capture the number of elements in its list.
\end{itemize}

\paragraph{Feature Vector Consistency}
A key aspect of our implementation is ensuring that every node, regardless of its type, produces a feature vector of the same length. Features that are not applicable to a given node are assigned a neutral default value. For example, the `Levenshtein Distance' feature is only computed for \textttnew{Identifier} nodes; for all other node types (e.g., \textttnew{Select}, \textttnew{Join}), it defaults to a high constant value representing ``no close match.'' Likewise, boolean features default to \textttnew{0}. This ensures that the GBDT model receives a consistently shaped input for any node in any given SQL query.

\section{Experiments}
\label{experiments}

\input{table2}

For both the dev slice of the BIRD dataset \cite{li2023can} and for SynSQL-2.5M \cite{li2025omnisql}, a very large synthetic dataset created for training the Text-to-SQL model OmniSQL \citep{li2023can}, we have generated SQL queries using the OmniSQL-7B model, which as of the writing of this paper is number 29 in the BIRD-SQL leaderboard. We chose this model because it is open-source, its close to SoTA performance, and its smaller size for running experiments efficiently.

Both the BIRD dev dataset and SynSQL-2.5M have multiple databases, each one with multiple tables: BIRD dev has 11 (and a total of 1534 queries), and for SynSQL-2.5M we used only 453 databases (and a total of 67,901 queries) rather than making generations for the entire set. We tested both in-database performance, and cross-database performance. In the first experiment we report in-database performance on BIRD dev: we sliced the queries related to each database into Train (80\%) and Test (20\%); and we trained a single model across all of the Train queries across all databases, testing on the Test queries across all databases. For the model we used LightGBM \citep{ke2017lightgbm} with parameters \textttnew{n\_estimators=100, learning\_rate=0.05}.
The results are in Table \ref{withinexperiment}. We can see that our approach is highly effective in detecting whether generated nodes are correct, with the following results on common node types: 69.59\% AUC on \textttnew{Literal}, 63.91\% AUC on \textttnew{Identifier}, and 61.48\% on \textttnew{TableAlias}. We can also see that this approach is far superior to log probabilities, whose AUC is similar to random chance, suggesting that the principle ``very uncertain generations should be less likely to be correct'' articulated in \citealp{kuhn2023semantic} can be significantly improved upon in the case of structured output.

Our second experiment is across databases. We split the BIRD dev databases into Train Databases and Test Databases (california schools, card\_games, toxicology). We trained on all of the queries in the train databases in BIRD dev, and tested on all of the queries in the test databases of BIRD dev. In the third experiment we trained on the entire 453 databases on which we generated queries in SynSQL-2.5M, and tested on the same test databases in BIRD dev as in the previous experiment, for a fair comparison.

As shown by results in Table \ref{acrossdatabasenexperiment}, both are superior to log probability and are reasonably good on \textttnew{Literal}. However, we observe that the more that the training data is in-distribution the better the model performs: Experiment 1 trained on the same databases it tested on; Experiment 2 trained on different databases from test, but from the same dataset of queries; and Experiment 3 was across datasets and performed the worst. This suggests that while in principle it might be possible to make a foundation model for node-level uncertainty, it is currently still best to train in-distribution.

\section*{Conclusions}
We introduced a semantically grounded framework for classifying correctness of LLM-generated SQL at the node level. Our approach contributes the first semantically aware labeling algorithm, a rich set of schema-aware features for node classification, and demonstrates substantial gains over log-probability baselines across multiple datasets. Together, these advances establish a practical and interpretable foundation for fine-grained uncertainty estimation in Text-to-SQL generation.
\section*{Limitations}
The ground truth labeling procedure in Section \ref{ssec:ground_truth} was designed to reflect intuitive assessments of uncertainty and has shown strong alignment in practice. While the method offers a solid foundation for meaningful analysis, occasional divergence from intuition may arise due to the inherent ambiguity in defining node-level correctness. Finally, we view the integration of uncertainty estimation into agentic mechanisms that dynamically revise or guide SQL generation as a promising direction for future work.

\section{Bibliographical References}\label{sec:reference}

\bibliographystyle{lrec2026-natbib}
\bibliography{custom}

\newpage
\appendix
\section{Illustrative Node-Level Labeling Examples}
\label{app:examples}

This appendix presents 13 paired SQL comparisons (generated vs.\ gold) and the node-level labels our algorithm produces. Each example highlights a specific design choice: (i) contextual subtree alignment with early stop on recursive equivalence; (ii) blame suppression for structural containers; (iii) semantic equivalence (alias/qualification normalization; operator symmetry/anti-symmetry); and (iv) a final global, acontextual rescue pass that corrects greedy misalignments.

\subsection*{Ex.\ 1: Perfect Match}
\textbf{Generated:} \textttnew{SELECT name FROM people}\\
\textbf{Gold:} \textttnew{SELECT name FROM people}\\
\textbf{Nodes Labeled as Incorrect:} none.\\
\textbf{Why it works:} Early stop on full-subtree equivalence; no recursion needed.

\subsection*{Ex.\ 2: Content Error on Table Only}
\textbf{Generated:} \textttnew{SELECT name FROM artists}\\
\textbf{Gold:} \textttnew{SELECT name FROM artist}\\
\textbf{Nodes Labeled as Incorrect:} \textttnew{Table(artists)}, \textttnew{Identifier(artists)}.\\
\textbf{Why it works:} Content mismatch on base table identifier; column sibling \textttnew{name} remains correct (qualification semantics prevent cascaded blame). Without the global pass, the sibling \textttnew{Column(name)} can be provisionally over\-blamed due to local divergence; the global pass reclassifies it as correct.

\subsection*{Ex.\ 3: Content Error on Literal Only}
\textbf{Generated:} \textttnew{SELECT * FROM t WHERE a = 1}\\
\textbf{Gold:} \textttnew{SELECT * FROM t WHERE a = 2}\\
\textbf{Nodes Labeled as Incorrect:} \textttnew{Literal(1)} only.\\
\textbf{Why it works:} Containers (\textttnew{SELECT}, \textttnew{FROM}, \textttnew{WHERE}, and the comparison operator \textttnew{=}) are suppressed; the global pass rescues \textttnew{Star}, \textttnew{Table(t)}, and \textttnew{Column(a)}, leaving only \textttnew{Literal(1)} incorrect. Without the global pass, this case fails (siblings remain incorrectly blamed).

\subsection*{Ex.\ 4: Type Error on Operator Only}
\textbf{Generated:} \textttnew{SELECT * FROM t WHERE a > 1}\\
\textbf{Gold:} \textttnew{SELECT * FROM t WHERE a = 1}\\
\textbf{Nodes Labeled as Incorrect:} \textttnew{GT(a > 1)} only.\\
\textbf{Why it works:} Operator type differs; containers are suppressed; children are equivalent. Without the global pass, \textttnew{Star}, \textttnew{Table(t)}, and \textttnew{Column(a)} can be over\-blamed; the global pass rescues them, leaving only \textttnew{GT}.

\subsection*{Ex.\ 5: Structural Insertion Blamed (Clause + Children)}
\textbf{Generated:} \textttnew{SELECT * FROM t ORDER BY a}\\
\textbf{Gold:} \textttnew{SELECT * FROM t}\\
\textbf{Nodes Labeled as Incorrect:} \textttnew{Order(ORDER BY a)}, \textttnew{Ordered(a)}, \textttnew{Column(a)}, \textttnew{Identifier(a)}.\\
\textbf{Why it works:} Extra clause and its subtree are structural insertions and are blamed precisely; containers elsewhere are suppressed. Without the global pass, \textttnew{Star(*)} may be over\-blamed; the global pass rescues it, leaving only the \textttnew{ORDER BY} subtree.

\subsection*{Ex.\ 6: Structural Omission Not Blamed}
\textbf{Generated:} \textttnew{SELECT * FROM t}\\
\textbf{Gold:} \textttnew{SELECT * FROM t ORDER BY a}\\
\textbf{Nodes Labeled as Incorrect:} none (on the generated query).\\
\textbf{Why it works:} Omissions in the generated SQL are not penalized because node-level probabilities must attach to existing generated nodes. Without the global pass, \textttnew{Star}, \textttnew{Table(t)}, and \textttnew{Identifier(t)} can be incorrectly blamed; the global pass rescues them so omissions are not penalized.

\subsection*{Ex.\ 7: Symmetric Operator Equivalence}
\textbf{Generated:} \textttnew{SELECT * FROM t WHERE a = b}\\
\textbf{Gold:} \textttnew{SELECT * FROM t WHERE b = a}\\
\textbf{Nodes Labeled as Incorrect:} none.\\
\textbf{Why it works:} \textttnew{=} is symmetric; children can match in reverse order.

\subsection*{Ex.\ 8: Anti-Symmetric Operator Equivalence}
\textbf{Generated:} \textttnew{SELECT * FROM t WHERE a > b}\\
\textbf{Gold:} \textttnew{SELECT * FROM t WHERE b < a}\\
\textbf{Nodes Labeled as Incorrect:} none.\\
\textbf{Why it works:} \textttnew{>} and \textttnew{<} are anti-symmetric pairs; swapping children and flipping the operator preserves equivalence.

\subsection*{Ex.\ 9: Alias Names May Differ (Still Correct)}
\textbf{Generated:} \textttnew{SELECT x.name FROM artist AS x}\\
\textbf{Gold:} \textttnew{SELECT a.name FROM artist AS a}\\
\textbf{Nodes Labeled as Incorrect:} none.\\
\textbf{Why it works:} Alias wrappers are ignored and normalized via alias$\rightarrow$table mapping; qualified vs.\ qualified with different alias names is equivalent.

\subsection*{Ex.\ 10: Unused Alias Declaration Not an Error}
\textbf{Generated:} \textttnew{SELECT name FROM artist AS a}\\
\textbf{Gold:} \textttnew{SELECT name FROM artist}\\
\textbf{Nodes Labeled as Incorrect:} none.\\
\textbf{Why it works:} Alias presence/absence is normalized away when it does not change base-table resolution.

\subsection*{Ex.\ 11: Qualified vs.\ Unqualified Column (Unambiguous)}
\textbf{Generated:} \textttnew{SELECT a.name FROM artist AS a}\\
\textbf{Gold:} \textttnew{SELECT name FROM artist}\\
\textbf{Nodes Labeled as Incorrect:} none.\\
\textbf{Why it works:} Column identifiers match; either side unqualified and alias$\rightarrow$table normalization yield equivalence (no ambiguity in scope).

\subsection*{Ex.\ 12: Wrong Base Table is an Error}
\textbf{Generated:} \textttnew{SELECT name FROM albums AS a}\\
\textbf{Gold:} \textttnew{SELECT name FROM artist AS a}\\
\textbf{Nodes Labeled as Incorrect:} \textttnew{Table(albums)} and/or \textttnew{Identifier(albums)}.\\
\textbf{Why it works:} Base table differs; alias normalization cannot reconcile distinct base tables.

\subsection*{Ex.\ 13: Wrong Alias Used in Column is an Error}
\textbf{Generated:} \textttnew{SELECT b.name FROM artist AS a}\\
\textbf{Gold:} \textttnew{SELECT a.name FROM artist AS a}\\
\textbf{Nodes Labeled as Incorrect:} \textttnew{Column(b.name)} and/or \textttnew{Identifier(b)}.\\
\textbf{Why it works:} Qualifier \textttnew{b} does not resolve in scope to the correct base table; column reference is invalid under alias normalization.

\vspace{0.75em}
\noindent\textbf{What could go wrong (and why it does not here).}
\begin{itemize}
  \item \textbf{Greedy alignment cascades:} A local mismatch (e.g., a literal or operator) can transiently over\-blame siblings under contextual alignment. Blame suppression for structural containers, followed by the global rescue pass, reclassifies semantically equivalent nodes back to correct (empirically required in Ex.\ 2--6).
  \item \textbf{Strict structural equality brittleness:} Reordering children, operator symmetry, or qualified vs.\ unqualified columns would look “unequal.” Semantic equivalence rules (operator symmetry/anti\-symmetry, alias$\rightarrow$table normalization, unqualified\-vs\-qualified tolerance) prevent false errors (Ex.\ 7--11).
  \item \textbf{Alias noise:} Different alias names or unused aliases can trigger spurious errors. Ignoring alias wrappers and normalizing to base tables prevents this (Ex.\ 9--11).
  \item \textbf{Structural insertions vs.\ omissions:} Extra clauses are blamed precisely at the inserted subtree (Ex.\ 5), while omissions in the generated SQL are not blamed (Ex.\ 6), reflecting our node\-level probability objective.
\end{itemize}

\paragraph{Ablation Evidence.} Disabling the global (acontextual) pass causes 5 of the 13 examples (Ex.\ 2--6) to fail---precisely those relying on rescuing transiently over\-blamed siblings. This validates the necessity of the global pass for precise node\-level labels.

These examples mirror the unit tests in our repository and demonstrate how contextual alignment, semantic equivalence, container suppression, and global rescue work together: the first pass localizes and structures blame; the suppression step avoids noisy container blame; and the global pass restores recall by reclassifying semantically equivalent nodes that contextual alignment alone cannot certify.

\end{document}

%% file: table1.tex
\begin{table*}[htp]
  \centering
  \begin{tabular}{lllll}
    \hline
    \textbf{Node Type} & \textbf{ Prop.}          & \textbf{Prop. False} & \textbf{AUC (ours)}  & \textbf{AUC (Logprobs)}\\
    \hline
    All & 100\% & 13.07\%      &  \textbf{76.51\%}           &       49.07\%        \\
         Identifier         & 40.06\%                &  9.74\% & \textbf{63.91\%} & 51.42\%       \\
    Column       & 15.25\%           & 8.32\%                  & \textbf{56.92\%} & 49.29\%         \\
    Literal & 5.99\%     &   10.35\%    & \textbf{69.59\%} & 42.44\%                    \\
        Table     &   5.73\%  & 4.66\% & 48.25\% & \textbf{49.63}\% \\
        TableAlias     &   4.90\%  & 6.61\% & \textbf{61.48\%} & 54.18\% \\
    \hline
  \end{tabular}
  \caption{\label{withinexperiment}
    Results for the first experiment, where each of the databases in BIRD dev is split into training and test sets. The model is trained on all of the training sets and tested on the test sets across the databases. ``Prop.'' is how often this node occurs in the LLM-generated data; and ``Prop. False'' is the ratio of nodes that are deemed as False compared to ground truth (see Section \ref{ssec:ground_truth}). Here, AUC refers to area under the ROC curve.
  }
\end{table*}

%% file: table2.tex

\begin{table*}[htbp]
  \centering
  \begin{tabular}{llll}
    \hline
    \textbf{Node Type} &  \textbf{Trained on BIRD} & \textbf{Trained on SynSQL-2.5M}  & \textbf{Logprobs AUC}\\
    \hline
    All &  \textbf{69.46\%}      &     66.47\%       &       48.45\%        \\
         Identifier                    &  50.43\% & \textbf{56.14\%} & 48.63\%       \\
    Column               & 45.48\%                  & \textbf{51.72\%} & 46.79\%         \\
    Literal     &   \textbf{58.91\%}    & 57.93\% & 51.16\%                    \\
        Table     & 52.61\% & \textbf{52.83\%} & 46.91\% \\
        TableAlias     & 51.43\% & \textbf{67.67\%} & 42.53\% \\
    \hline
  \end{tabular}
  \caption{\label{acrossdatabasenexperiment}
    Results for the second and third experiments. In this table, the evaluation is done on the queries associated with the databases california schools, card\_games, and toxicology in the BIRD dev dataset. The column ``Trained on BIRD'' refers to a model trained on all of the queries in the BIRD dev dataset not associated with these 3 databases. The ``Trained on SynSQL-2.5M'' column refers to a model trained \textit{only} on queries from 453 databases from SynSQL-2.5M. AUC in this table refers to area under the ROC curve.
  }
\end{table*}